\documentclass[10pt,twocolumn]{article}
\usepackage[letterpaper,margin=0.68in]{geometry}
\usepackage{times}
\usepackage{microtype}
\usepackage{amsmath,amssymb}
\usepackage{booktabs}
\usepackage{graphicx}
\usepackage{tikz}
\usetikzlibrary{positioning}
\usepackage{enumitem}
\usepackage[hidelinks]{hyperref}
\usepackage{url}
\usepackage{xcolor}
\usepackage{xspace}
\setlist[itemize]{leftmargin=1.35em,itemsep=1pt,topsep=2pt}
\newcommand{\CE}{\mathrm{CE}}
\newcommand{\FDAA}{\textsc{FDAA}\xspace}
\title{\textbf{Beyond Routing: Decoupling Expert Dispatch and Aggregation in\\Sparse Mixture-of-Experts}}
\author{Zongfei Li}
\date{}

\begin{document}
\maketitle
\vspace{-1.2em}

\begin{abstract}
Sparse Mixture-of-Experts (MoE) routers conventionally use the same scores both to select experts and to weight their already-computed outputs. We ask whether these two roles---\emph{dispatch} and \emph{aggregation}---should be coupled. On pretrained OLMoE-1B-7B, we keep selected Top-8 expert IDs, expert computation, and total selected router mass fixed and intervene only on within-set aggregation. A structured oracle improves full-horizon cross-entropy by $0.0160\pm0.0039$ across three seeds; the router's top-scored expert is the counterfactual-best vertex only 17.2\% of the time, with router--utility Spearman 0.030. Downstream gradients expose useful commitment credit, but multiple forward-only credit predictors fail to generalize. We therefore train Fixed-Dispatch Adaptive Aggregation (\FDAA), a 301K-parameter post-compute head optimized directly with the language-modeling objective while freezing the backbone, router, and experts. On OLMoE, a layer-15 head improves fresh WikiText-103 test by $\Delta\CE=-0.1523\pm0.0031$ across three seeds. Under deterministic 50/50 WikiText+C4 training, frozen confirmatory evaluation on 1,024 fixed windows per corpus improves WikiText-103, C4, and held-out Penn Treebank in all three seeds. We then replicate the fixed-dispatch audit on DeepSeek-V2-Lite, which uses Top-6 routed experts plus shared experts. Best-vertex headroom remains significant on both WikiText ($-0.00867$) and C4 ($-0.00856$), while router Top1 identifies the best selected expert in only 12.5\% and 16.7\% of audited examples. In a one-seed mixed-domain replication, \FDAA improves locked WikiText test by $-0.01747$ and PTB test by $-0.00831$, significantly outperforming learned static calibration on both; C4 is statistically neutral. Together, these results support a cross-architecture distinction between expert selection and expert commitment, while showing that the magnitude and transferability of learned commitment remain domain- and architecture-dependent.
\end{abstract}

\section{Introduction}
Sparse Mixture-of-Experts (MoE) models scale parameter count without proportionally scaling per-token computation by routing each token to a small subset of experts \cite{shazeer2017,lepikhin2020,fedus2021}. Modern language-model MoEs such as Mixtral, DeepSeekMoE, and OLMoE use a learned router to select active experts and combine their outputs \cite{jiang2024,dai2024,muennighoff2024}. This design implicitly assigns the router two responsibilities: \textbf{dispatch}, which decides which experts are executed, and \textbf{aggregation}, which decides how much each selected expert contributes after it has already been computed.

These responsibilities need not be identical. A score may be sufficient to rank a useful candidate set while being a poor estimate of each selected expert's downstream causal value. This distinction matters because changing dispatch is expensive and operationally intrusive: it changes which expert kernels execute, communication patterns, capacity allocation, and often load-balancing behavior. By contrast, once the selected expert outputs are already available, changing only their aggregation weights can preserve the sparse execution pattern exactly.

We study this possibility through \emph{dispatch--aggregation decoupling}. Our central experimental constraint is deliberately strict: for every intervention, we keep the router-selected Top-$K$ expert IDs fixed, compute exactly the same selected experts, and preserve the total selected Top-$K$ router mass. Only the relative weights over the already-computed selected expert outputs may change. This creates a clean causal question: \emph{conditional on dispatch, are the router's aggregation weights already optimal?}

Our experiments reveal a consistent selection--commitment mismatch on OLMoE-1B-7B. Structured counterfactual reweighting exhibits nontrivial full-horizon headroom across seeds. The router's highest-scored selected expert is rarely the best counterfactual commitment choice, and its within-set score ranking has near-zero correlation with exact utility. A first-order downstream gradient, however, strongly predicts which selected expert should receive more commitment. This suggests that the missing signal is not expert capacity but downstream credit assignment.

A natural next step is to distill this credit into a forward-only controller. We systematically test that route and find a useful negative result: direct counterfactual utility distillation, gradient-field prediction, and a task-aware 8-D credit-subspace predictor all fail to generalize. The task-aware representation itself reconstructs gradient utility essentially perfectly, yet even leaked full-suffix hidden states do not make the credit reliably predictable. This motivates a different strategy: rather than requiring a frozen representation to already encode full-horizon credit, let the aggregation policy adapt directly under the true language-modeling objective.

We introduce \textbf{Fixed-Dispatch Adaptive Aggregation (\FDAA)}, a lightweight token-adaptive head applied after selected experts have been computed. The pretrained model, router, and experts remain frozen; only the aggregation head is optimized. The head starts exactly at the native router aggregation through zero initialization and preserves selected Top-$K$ mass by construction. Locked and fresh OLMoE experiments show large, statistically stable improvements; token-shuffle and learned static-calibration controls establish that the gain is genuinely token-adaptive. A frozen confirmatory evaluation further shows that one mixed-domain OLMoE head improves WikiText-103 test, C4 validation, and held-out PTB test in all three optimization seeds. A second-backbone audit on DeepSeek-V2-Lite replicates the fixed-dispatch headroom and near-chance router ranking under a different Top-6-plus-shared-expert architecture; a one-seed \FDAA replication significantly improves WikiText and PTB while remaining neutral on C4.

Our contributions are:
\begin{itemize}
\item \textbf{A causal decomposition of sparse MoE routing.} We separate expert dispatch from post-dispatch aggregation and audit the latter while holding selected experts and computation fixed.
\item \textbf{Cross-architecture evidence of selection--commitment mismatch.} OLMoE Top-8 and DeepSeek-V2-Lite Top-6 both contain substantially better within-set commitment choices than their native router ranking expresses.
\item \textbf{A credit-assignment diagnosis.} On OLMoE, downstream first-order gradients strongly predict counterfactual commitment utility, while multiple forward-only post-hoc predictors fail even under favorable representations and future leakage.
\item \textbf{A simple trainable remedy.} \FDAA directly learns token-adaptive aggregation under the language-model objective with about 301K trainable parameters, no expert rerouting, and no additional expert execution.
\item \textbf{Strong controls, transfer, and a second-backbone replication.} Three-seed OLMoE results beat static calibration and token shuffle across fresh and confirmatory evaluations. A one-seed DeepSeek-V2-Lite replication significantly improves locked WikiText and PTB and beats learned static calibration on both, while C4 remains neutral.
\end{itemize}

\section{Dispatch and Aggregation Are Distinct Decisions}
\subsection{Standard sparse MoE}
Let $h_t\in\mathbb{R}^d$ be the input to an MoE block for token $t$. A router produces logits over $N$ experts and selects a set $S_t$ of $K$ experts. For selected expert outputs $u_{t,j}=E_j(h_t)$ and router weights $w_{t,j}\ge0$, the native MoE output is
\begin{equation}
 m_t=\sum_{j\in S_t} w_{t,j}u_{t,j}.
\end{equation}
The router scores therefore affect both $S_t$ and the coefficients in Eq.~1. We define the total selected router mass
\begin{equation}
 M_t=\sum_{j\in S_t} w_{t,j}.
\end{equation}
Our interventions keep $S_t$, every $u_{t,j}$, and $M_t$ fixed. A counterfactual aggregation uses alternative nonnegative weights $w'_{t,j}$ satisfying
\begin{equation}
 \sum_{j\in S_t}w'_{t,j}=M_t.
\end{equation}
No additional expert is evaluated. Thus any change in downstream loss is attributable to commitment within the selected set, not to rerouting.

\subsection{Why this decomposition is useful}
Recent counterfactual routing work asks whether a different equal-compute route would have been better \cite{yoon2026}. Our question is orthogonal: even when the selected set is held fixed, are the selected experts weighted well? Likewise, recent aggregation architectures add structural interaction or message passing among selected experts \cite{feng2026,kulibaba2026}. We instead retain the standard weighted-sum operator and isolate only the scalar commitment weights. This restricted intervention provides a lower-level diagnostic and can be attached to a pretrained MoE without altering expert execution.

\section{Diagnosing Selection--Commitment Mismatch}
\subsection{Fixed-dispatch aggregation oracle}
For OLMoE-1B-7B-0924, we audit layers $\{3,7,11,15\}$ and use a structured action family containing the native baseline, uniform aggregation, temperatures 0.5 and 2.0, eight 50\% mixtures toward individual selected-expert slots, and eight vertices that place the full selected router mass on one already-selected expert. Every action preserves selected IDs and total Top-$K$ mass.

Across three seeds, the locked full-horizon structured oracle obtains $\Delta\CE=-0.02015,-0.01234,-0.01548$, or $-0.01599\pm0.00393$. A higher-powered decomposition audit ($n=128$) shows that a globally fixed action selected on development data does not improve locked loss ($+0.00122$), whereas the token-adaptive structured oracle reaches $-0.02172$. The vertex family alone reaches $-0.02011$, explaining 92.6\% of structured-oracle gain (Table~\ref{tab:diag}). Thus most available headroom can be exposed by changing relative commitment among already-selected experts rather than by inventing a more complex continuous mixture.

\begin{table}[t]
\centering
\caption{Full-horizon fixed-dispatch diagnostic on OLMoE ($n=128$, seed 1337 unless noted). Negative $\Delta\CE$ is better.}
\label{tab:diag}
\scriptsize
\resizebox{\columnwidth}{!}{%
\begin{tabular}{lrl}
\toprule
Quantity & Value & Interpretation\\
\midrule
3-seed structured oracle & $-0.01599\pm0.00393$ & stable headroom\\
Dev-fixed action & $+0.00122$ & global rule fails\\
Structured oracle & $-0.02172$ & adaptive lower bound\\
Vertex oracle & $-0.02011$ & 92.6\% capture\\
Router Top1=best vertex & 17.2\% & weak rank\\
Mean rank best vertex & 4.09/8 & near middle\\
Router--utility Spearman & 0.030 & near zero\\
\bottomrule
\end{tabular}%
}
\end{table}

\subsection{The router selects a useful set but poorly ranks commitment}
The native router's top-scored selected expert is the exact best vertex in only 17.2\% of locked full-horizon samples. The mean rank of the best expert is 4.09 among eight selected experts, and the mean within-Top-8 Spearman correlation between router score and exact vertex utility is 0.030. These results motivate a distinction between candidate-set quality and commitment quality: the useful expert may already be selected, yet the router score provides little guidance about how strongly to use it after selection.

\subsection{Downstream gradients expose commitment credit}
For target MoE output $m_t$, let $g_t=\partial L/\partial m_t$ be the downstream gradient of a horizon loss. For a fixed-dispatch vertex that replaces $m_t$ by $M_tu_{t,j}$, the first-order predicted improvement is
\begin{equation}
 \widehat{G}_{t,j}=-g_t^\top(M_tu_{t,j}-m_t).
\end{equation}
Using isolated batch-1 runs to avoid sparse-MoE co-batch numerical coupling, the full-horizon gradient utility obtains mean within-Top-8 Spearman 0.595, pairwise accuracy 0.749, and exact-best Top1 accuracy 50.4\%, versus 12.5\% Top1 for the router. Selecting the gradient-best vertex reduces CE by 0.01073 and captures 53.1\% of exact vertex-oracle gain. A development-selected conservative threshold reduces harm to 2.34\% of samples while retaining $\Delta\CE=-0.00908$ on locked data. This result is diagnostic rather than deployable: it requires backward information from the target loss.

\subsection{Negative result: post-hoc credit prediction is not enough}
A direct 810K-parameter counterfactual reranker trained from exact labels fails to generalize (locked utility Spearman 0.036; raw policy $\Delta\CE=+0.00275$). Gradient-field prediction likewise remains close to random. To rule out the possibility that predicting the full hidden-size gradient is simply the wrong representation, we build a sample-specific task-aware credit subspace from the eight expert displacement vectors. Its pseudoinverse reconstruction is lossless for gradient-utility ranking on locked data, yet the best deployable predictor reaches only Spearman 0.089. Even a deliberately leaked predictor given the complete realized future suffix obtains Spearman 0.008. We therefore optimize aggregation directly under the true LM objective.

\section{Fixed-Dispatch Adaptive Aggregation}
\subsection{Parameterization}
\FDAA receives the current token representation, the already-computed selected expert outputs, the native selected router weights, selected expert IDs, and a layer embedding. For each selected slot it predicts a residual commitment score $s_{t,j}$. Let
\begin{equation}
 p_{t,j}=\frac{w_{t,j}}{M_t}.
\end{equation}
We define
\begin{equation}
 q_{t,j}=\frac{p_{t,j}\exp(s_{t,j})}{\sum_{\ell\in S_t}p_{t,\ell}\exp(s_{t,\ell})},\qquad
 w'_{t,j}=M_tq_{t,j}.
\end{equation}
By construction, $\sum_jw'_{t,j}=M_t$. The final score layer is zero initialized, so $s_{t,j}=0$ at initialization and the modified aggregation reproduces the native model up to numerical precision.

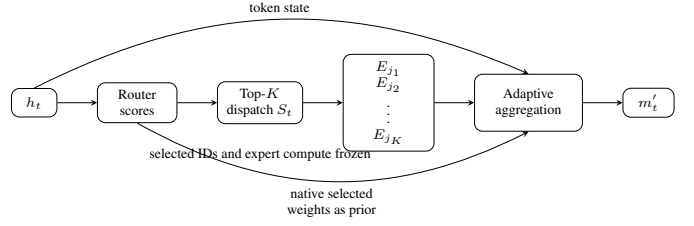
\begin{figure}[t]
\centering
\resizebox{\columnwidth}{!}{%
\begin{tikzpicture}[>=stealth, node distance=5mm, every node/.style={font=\scriptsize}]
\node (h) [draw, rounded corners, minimum width=8mm, minimum height=5mm] {$h_t$};
\node (r) [draw, rounded corners, right=7mm of h, minimum width=14mm, minimum height=7mm, align=center] {Router\\scores};
\node (d) [draw, rounded corners, right=7mm of r, minimum width=15mm, minimum height=7mm, align=center] {Top-$K$\\dispatch $S_t$};
\node (e) [draw, rounded corners, right=7mm of d, minimum width=16mm, minimum height=14mm, align=center] {$E_{j_1}$\\$E_{j_2}$\\$\vdots$\\$E_{j_K}$};
\node (a) [draw, rounded corners, right=7mm of e, minimum width=19mm, minimum height=10mm, align=center] {Adaptive\\aggregation};
\node (m) [draw, rounded corners, right=7mm of a, minimum width=9mm, minimum height=5mm] {$m'_t$};
\draw[->] (h)--(r); \draw[->] (r)--(d); \draw[->] (d)--(e); \draw[->] (e)--(a); \draw[->] (a)--(m);
\draw[->,bend right=28] (r.south) to node[below,align=center] {native selected\\weights as prior} (a.south);
\draw[->,bend left=25] (h.north) to node[above] {token state} (a.north);
\node[below=3mm of d, align=center] {selected IDs and expert compute frozen};
\end{tikzpicture}%
}
\caption{Dispatch--aggregation decoupling. The original router still selects the Top-$K$ experts, and exactly those experts are executed. \FDAA only reweights their already-computed outputs while preserving total selected router mass.}
\label{fig:diagram}
\end{figure}

The head is permutation-equivariant over selected expert slots. It projects the token state and expert outputs into compact features, includes expert-identity and layer embeddings, and predicts one scalar residual score per selected expert. The Stage-1 configuration contains 301,313 trainable parameters, negligible relative to the 7B-parameter checkpoint. The pretrained model, router, and expert parameters have gradients disabled.

\subsection{Training objective}
We optimize the ordinary next-token language-modeling cross-entropy through the modified aggregation output, with a small KL penalty from $q_t$ to the normalized native router weights $p_t$ and an $\ell_2$ penalty on residual scores:
\begin{equation}
 L=L_{\mathrm{LM}}+\lambda_{\mathrm{KL}}D_{\mathrm{KL}}(q_t\Vert p_t)+\lambda_s\lVert s_t\rVert_2^2.
\end{equation}
The regularizers are stability terms, not teacher signals. No counterfactual labels or gradient-utility labels are required.

\subsection{Stage-1 and layer localization}
We initially share a layer-conditioned head across layers $\{3,7,11,15\}$ and modify one target layer per training microbatch. All four layers improve, but layer 15 dominates: on locked data layer 15 yields $\Delta\CE=-0.1926$, whereas enabling all four learned interventions gives $-0.1934$. A later three-seed fresh-test audit confirms that all four layers together do not outperform layer 15 alone. We therefore localize subsequent multi-domain training to layer 15.

\section{Experimental Setup}
\paragraph{Primary backbone.} We use \texttt{allenai/OLMoE-1B-7B-0924}, a 16-layer sparse MoE with 64 experts and Top-8 routing in our implementation. OLMoE has 7B total parameters and roughly 1B active parameters per token \cite{muennighoff2024}. Unless otherwise stated, sequence length is 96 and evaluation interventions are run independently with batch size 1.

\paragraph{Second backbone.} We additionally replicate on \texttt{deepseek-ai/DeepSeek-V2-Lite} \cite{deepseekv2}. In our implementation it has 27 layers, MoE blocks in layers 1--26, 64 routed experts with Top-6 routing, and two shared experts. We target the final MoE layer (26). The DeepSeek head has 301,489 trainable parameters; the learned static calibrator has 71.

\paragraph{Strict fixed-dispatch protocol.} For OLMoE, selected Top-8 IDs are taken from the native router; for DeepSeek-V2-Lite, selected routed Top-6 IDs are fixed while the shared-expert branch is left untouched. In both cases, exactly the native selected routed experts are executed and their total selected router mass is preserved. Zero-initialized heads reproduce native logits and CE exactly (up to numerical roundoff), and selected-ID mismatch is required to be zero before nontrivial reweighting.

\paragraph{Data and training.} Stage-1 OLMoE training uses WikiText-103 train. The validation corpus is physically split for development and locked evaluation; selected heads are subsequently audited on untouched WikiText-103 test. For the multi-domain extension, we train separate WikiText-only, C4-only, and deterministic 50/50 WikiText+C4 layer-15 heads. Checkpoint selection uses WikiText validation and held-out C4 training-stream tokens. WikiText-103 test and C4 validation are excluded from checkpoint selection, and Penn Treebank is never used for training or selection. For the DeepSeek replication, we use 1.2M training tokens from each of WikiText and C4 and select checkpoints using an equal-domain mean normalized-CE objective on non-overlapping development windows. The one-seed locked evaluation uses 128 length-96 windows per corpus from WikiText-103 test, C4 validation, and PTB test.

\paragraph{Frozen confirmatory evaluation.} After all mixed-domain checkpoints are frozen, we evaluate seeds 1337, 2027, and 4049 on 1,024 fixed, non-overlapping length-96 windows from each of WikiText-103 test, C4 validation, and PTB test. The same windows are used across optimization seeds within each corpus. We use batch size 1 and 10,000 bootstrap resamples. For the primary cross-seed confidence interval, we first average the three seed deltas within each evaluation window and then bootstrap over windows; we do not treat seed$\times$window cells as independent observations.

\paragraph{Baselines and controls.} We compare against native router aggregation, uniform weights, temperature transforms, a learned static layer-by-expert residual-score calibrator, specialist heads trained on one domain, and token-shuffled adaptive weights. Token shuffle applies a policy derived from another position in the same sequence while preserving the current token's selected Top-$K$ mass; it is a direct negative control for token-specific adaptivity.

\paragraph{Statistics.} We report mean CE difference relative to independent native forwards; negative values are improvements. Confidence intervals are nonparametric bootstrap 95\% intervals. For earlier three-seed experiments we additionally report mean $\pm$ sample standard deviation across optimization seeds. Development-selected policies are always evaluated on locked or fresh splits after selection.

\section{Results}
\subsection{Direct LM training succeeds where post-hoc predictors fail}
Table~\ref{tab:stage1} reports the Stage-1 locked evaluation. The learned aggregation head improves pooled single-layer CE by $-0.07626$ with a bootstrap interval entirely below zero. Uniform and temperature-2 calibration provide much smaller gains, while temperature-0.5 is harmful. On exactly paired sample/layer cells, the learned head beats the strongest fixed transform by $-0.06250$ CE.

\begin{table}[t]
\centering
\caption{Stage-1 locked WikiText evaluation (seed 1337). Negative is better.}
\label{tab:stage1}
\small
\begin{tabular}{lrr}
\toprule
Policy & Mean $\Delta\CE$ & 95\% CI\\
\midrule
\FDAA pooled & $-0.07626$ & $[-0.08971,-0.06415]$\\
Uniform & $-0.01376$ & $[-0.02070,-0.00724]$\\
Temp. 2.0 & $-0.00693$ & $[-0.00988,-0.00427]$\\
Temp. 0.5 & $+0.02406$ & $[+0.02117,+0.02731]$\\
\FDAA layer 15 & $-0.19260$ & $[-0.23665,-0.15604]$\\
\FDAA all 4 & $-0.19341$ & $[-0.23796,-0.15646]$\\
Learned $-$ best fixed & $-0.06250$ & $[-0.07270,-0.05369]$\\
\bottomrule
\end{tabular}
\end{table}

The comparison is especially informative because post-hoc credit models saw explicit counterfactual or gradient supervision yet failed, whereas direct LM training succeeds without an auxiliary teacher. The optimization objective can shape the aggregation policy and its internal features jointly rather than asking a frozen representation to expose downstream credit in an easily decodable form.

\subsection{Token adaptivity replicates across three seeds on fresh WikiText}
We evaluate three independently trained Stage-1 heads on fresh WikiText-103 test windows. Layer 15 alone yields $\Delta\CE=-0.15228\pm0.00313$ across seeds (Table~\ref{tab:fresh}). It exceeds a separately optimized static layer-by-expert calibrator by $-0.12099\pm0.00313$ and a token-shuffled version of the learned policy by $-0.16375\pm0.00488$. The all-four-layer intervention is slightly worse than layer 15 alone, so subsequent mixed-domain training uses only layer 15.

\begin{table}[t]
\centering
\caption{Fresh WikiText-103 test, three optimization seeds.}
\label{tab:fresh}
\small
\begin{tabular}{lr}
\toprule
Quantity & $\Delta\CE$\\
\midrule
Layer-15 adaptive head & $-0.15228\pm0.00313$\\
All four audited layers & $-0.14794\pm0.00470$\\
Layer15 $-$ static calibrator & $-0.12099\pm0.00313$\\
Layer15 $-$ token shuffle & $-0.16375\pm0.00488$\\
All4 $-$ Layer15 & $+0.00434\pm0.00269$\\
\bottomrule
\end{tabular}
\end{table}

\subsection{Single-domain transfer is asymmetric}
WikiText-only heads do not zero-shot generalize to C4: across the same three seeds, layer-15 CE changes by $+0.01695\pm0.00090$. Interestingly, the token-aware policy still outperforms its token-shuffled counterpart on C4, suggesting that it has learned meaningful token-conditioned behavior but one adapted to the WikiText training distribution. This failure separates token adaptivity from domain generality and motivates broader training.

\subsection{Mixed-domain training yields confirmatory cross-domain gains}
We train a layer-15 head with deterministic 50/50 WikiText+C4 microbatches using the same hyperparameters across seeds. We then freeze all checkpoints and run the confirmatory protocol described above. Table~\ref{tab:confirm} summarizes the window-mean-across-seeds analysis.

On WikiText-103 test, the mixed head improves native CE by $-0.19367$ with 95\% CI $[-0.20609,-0.18143]$. On C4 validation, it improves CE by $-0.00487$ with CI $[-0.00581,-0.00390]$. Crucially, the C4 result is now individually significant for all three optimization seeds: $-0.00476$, $-0.00475$, and $-0.00508$. The mixed head also beats the static calibrator by $-0.00625$ and token shuffle by $-0.02251$ on C4, with both paired intervals strictly below zero.

Held-out PTB provides a second transfer test that is completely absent from training and checkpoint selection. All three seeds are evaluated on the same PTB \emph{test} split and improve native CE: $-0.05982$, $-0.06365$, and $-0.05840$. The cross-seed window mean is $-0.06063$ with CI $[-0.06302,-0.05826]$ and a 95.1\% win fraction. The mixed head beats static calibration by $-0.05369$, token shuffle by $-0.06001$, and temperature-2 by $-0.04958$ on PTB.

The magnitude remains domain-sensitive: C4 gains are substantially smaller than WikiText or PTB gains. We therefore do not claim domain-invariant commitment. The stronger conclusion is that broader training converts the original WikiText-to-C4 failure into a small but repeatable improvement while preserving robust token-conditioned advantages over static and shuffled controls.

\begin{table*}[t]
\centering
\caption{Frozen three-seed confirmatory evaluation. Each corpus uses 1,024 fixed non-overlapping windows shared across seeds. Policy rows show the mean of the per-window seed average; brackets are 95\% bootstrap CIs over windows. Negative is better.}
\label{tab:confirm}
\small
\resizebox{\textwidth}{!}{%
\begin{tabular}{lccc}
\toprule
Quantity & WikiText-103 test & C4 validation & PTB test\\
\midrule
Mixed shared & $-0.19367\;[-0.20609,-0.18143]$ & $-0.00487\;[-0.00581,-0.00390]$ & $-0.06063\;[-0.06302,-0.05826]$\\
Static mixed & $-0.04328\;[-0.04603,-0.04065]$ & $+0.00138\;[+0.00102,+0.00174]$ & $-0.00694\;[-0.00742,-0.00647]$\\
Token-shuffled mixed & $+0.00441\;[+0.00045,+0.00820]$ & $+0.01765\;[+0.01616,+0.01917]$ & $-0.00061\;[-0.00209,+0.00087]$\\
Temp. 2.0 & $-0.01645\;[-0.01816,-0.01472]$ & $+0.00385\;[+0.00309,+0.00461]$ & $-0.01105\;[-0.01187,-0.01022]$\\
\midrule
Mixed $-$ static & $-0.15039\;[-0.16074,-0.14065]$ & $-0.00625\;[-0.00712,-0.00534]$ & $-0.05369\;[-0.05599,-0.05132]$\\
Mixed $-$ shuffle & $-0.19808\;[-0.20940,-0.18748]$ & $-0.02251\;[-0.02425,-0.02087]$ & $-0.06001\;[-0.06249,-0.05747]$\\
Mixed $-$ Temp. 2.0 & $-0.17722\;[-0.18884,-0.16635]$ & $-0.00872\;[-0.00989,-0.00755]$ & $-0.04958\;[-0.05187,-0.04728]$\\
\midrule
Individual sig. seeds vs native & 3/3 & 3/3 & 3/3\\
Individual sig. seeds vs static & 3/3 & 3/3 & 3/3\\
Individual sig. seeds vs shuffle & 3/3 & 3/3 & 3/3\\
\bottomrule
\end{tabular}%
}
\end{table*}

\subsection{Second-backbone replication on DeepSeek-V2-Lite}
We next test whether the selection--commitment mismatch is specific to OLMoE. DeepSeek-V2-Lite differs materially in routing structure: at the audited final MoE layer it uses Top-6 routed experts together with two shared experts. We keep the routed Top-6 IDs, routed expert computation, and total routed mass fixed and leave the shared-expert branch unchanged. Sanity checks give zero zero-initialized logit and CE mismatch, zero mass error, and zero selected-ID mismatch.

The fixed-dispatch diagnostic replicates. On 24 WikiText development examples, the best routed-expert vertex improves CE by $-0.00867$ with 95\% CI $[-0.01426,-0.00449]$, while the router's Top1 selected expert is the best vertex only 12.5\% of the time. On 24 C4 development examples, the best vertex improves CE by $-0.00856$ with CI $[-0.01413,-0.00411]$, and router Top1 matches the best vertex in 16.7\% of examples---exactly the Top-6 chance rate. Committing to the router Top1 on C4 is itself harmful on average ($+0.00411$, CI $[+0.00146,+0.00738]$). Thus the core mismatch is not unique to OLMoE.

A one-seed mixed-domain \FDAA replication also yields positive but domain-dependent learned results (Table~\ref{tab:deepseek}). On locked WikiText-103 test, \FDAA improves native CE by $-0.01747$ (CI $[-0.02061,-0.01440]$) and beats the learned static calibrator by $-0.01289$ (CI $[-0.01537,-0.01029]$). On held-out PTB test, it improves native CE by $-0.00831$ (CI $[-0.01097,-0.00587]$) and beats static calibration by $-0.00377$ (CI $[-0.00606,-0.00171]$). C4 is statistically neutral: \FDAA changes CE by $+0.00011$ with CI $[-0.00086,+0.00106]$, and its paired difference from static is also indistinguishable from zero. We therefore treat this as a clear cross-architecture replication of the diagnostic mismatch and a \emph{partial} replication of the trainable remedy, not as evidence of domain-uniform transfer.

\begin{table*}[t]
\centering
\caption{One-seed DeepSeek-V2-Lite replication. Diagnostic oracle rows use 24 development examples per domain; learned-policy rows use 128 locked windows per evaluation corpus. Negative $\Delta\CE$ is better.}
\label{tab:deepseek}
\small
\resizebox{\textwidth}{!}{%
\begin{tabular}{lccc}
\toprule
Quantity & WikiText & C4 & PTB test\\
\midrule
Best-vertex oracle (dev) & $-0.00867\;[-0.01426,-0.00449]$ & $-0.00856\;[-0.01413,-0.00411]$ & --\\
Router Top1 = best vertex & 12.5\% & 16.7\% & --\\
\midrule
\FDAA locked & $-0.01747\;[-0.02061,-0.01440]$ & $+0.00011\;[-0.00086,+0.00106]$ & $-0.00831\;[-0.01097,-0.00587]$\\
Static locked & $-0.00458\;[-0.00571,-0.00341]$ & $+0.00003\;[-0.00071,+0.00075]$ & $-0.00455\;[-0.00551,-0.00358]$\\
\FDAA $-$ static & $-0.01289\;[-0.01537,-0.01029]$ & $+0.00008\;[-0.00095,+0.00113]$ & $-0.00377\;[-0.00606,-0.00171]$\\
\FDAA $-$ shuffle & $-0.11947\;[-0.15189,-0.09305]$ & $-0.03621\;[-0.04702,-0.02553]$ & $-0.33604\;[-0.35933,-0.31415]$\\
\bottomrule
\end{tabular}%
}
\end{table*}

\subsection{What information does the head use?}
Inference-time feature masking on the three WikiText-trained heads provides an additional diagnostic. The full layer-15 head obtains $-0.1523\pm0.0031$ on fresh WikiText test. Removing expert outputs leaves only $-0.01184\pm0.00328$, while using router and expert-ID information alone leaves $-0.00377\pm0.00117$; removing the token hidden state collapses the measured gain to zero. Although these are inference-time masks rather than retrained ablations, they suggest that the large gain depends on combining token state with post-compute expert information rather than memorizing static expert identities.

\section{Discussion}
\paragraph{Selection quality is not commitment quality.} Our counterfactual results do not imply that the router selected a bad expert set. In fact, the strong vertex oracle suggests the opposite: useful alternatives are often already present among the Top-8. The failure lies in mapping selected candidates to commitment weights. This distinction helps reconcile strong expert specialization with poor within-set router--utility correlation.

\paragraph{Why direct LM training beats credit distillation.} The gradient audit shows that the desired direction is available to backpropagation, but predictor audits show that full-horizon credit is not easily recoverable from the frozen forward state. Direct LM training avoids this representational bottleneck: the same objective that defines downstream utility is allowed to reshape the aggregation head's features.

\paragraph{The strongest effect is late-layer.} The final audited MoE layer accounts for most of the learned gain, and adding earlier learned interventions provides no extra benefit on fresh test. This is favorable for deployment because a useful version may require only one post-compute module. It also raises a mechanistic question: late layers may have less opportunity for subsequent blocks to repair misweighted expert outputs.

\paragraph{Domain breadth matters, but commitment remains domain-sensitive.} WikiText-only OLMoE training produces a highly effective in-domain policy that harms C4. Mixed training reverses that C4 failure and produces statistically significant improvements in all three seeds under the frozen confirmatory protocol. DeepSeek-V2-Lite sharpens the same conclusion from another angle: a mixed head significantly improves WikiText and PTB but is neutral on C4. Commitment is therefore transferable, but neither its magnitude nor its optimal policy should be assumed domain-invariant.

\paragraph{The mismatch generalizes more cleanly than the learned gain.} The second-backbone diagnostic is qualitatively striking: significant fixed-dispatch headroom persists under a Top-6 router with shared experts, while router Top1 is only chance-level at identifying the best routed-expert vertex. This strengthens the claim that selection and commitment are distinct problems across pretrained MoE designs. At the same time, the one-seed DeepSeek learned result is weaker and more domain-dependent than the OLMoE result. We view the cross-architecture replication of the \emph{phenomenon} as stronger than the current cross-architecture replication of the \emph{remedy}.

\section{Related Work}
\paragraph{Sparse MoE routing.} Sparsely gated MoEs use conditional computation to increase model capacity at roughly fixed active compute \cite{shazeer2017}. GShard and Switch Transformer developed scalable Top-$K$ and Top-1 routing systems \cite{lepikhin2020,fedus2021}; more recent language models such as Mixtral, DeepSeekMoE, and OLMoE demonstrate sparse expert models at modern scale \cite{jiang2024,dai2024,muennighoff2024}.

\paragraph{Is the router choosing the right experts?} Dense Backpropagation improves sparse-MoE router training by providing denser gradient information while retaining sparse activation \cite{panda2025}. Counterfactual routing analysis finds lower-loss equal-compute routes inside frozen pretrained MoEs and studies router-reachable misallocation across multiple backbones \cite{yoon2026}. These works primarily concern which experts execute. Our study conditions on the executed set and asks whether the relative aggregation weights over already-computed experts are appropriate.

\paragraph{Beyond weighted-sum aggregation.} DAG-MoE treats aggregation structure as a scaling axis and learns structured composition among selected experts \cite{feng2026}. SDG-MoE adds signed interaction and message passing before final aggregation \cite{kulibaba2026}. Our work is complementary but intentionally more restrictive: we preserve the ordinary weighted-sum form, do not change selected expert representations through inter-expert communication, and isolate the effect of scalar post-dispatch commitment weights on a frozen pretrained MoE.

\section{Limitations and Planned Revision}
The current evidence has several limitations. First, the strongest confirmatory evidence remains on OLMoE: three optimization seeds and 1,024 fixed windows per corpus. DeepSeek-V2-Lite provides a second-backbone replication of fixed-dispatch headroom and router mismatch, but its learned-policy result currently uses one optimization seed and 128 locked windows per corpus. The DeepSeek head significantly improves WikiText and PTB and beats static calibration on both, but it is neutral on C4; additional seeds and a larger locked evaluation are therefore needed before claiming fully replicated cross-architecture \FDAA performance. Second, our main metric is next-token cross-entropy on language-modeling corpora; downstream reasoning, knowledge, code, and instruction-following tasks remain to be tested. Third, although selected expert execution is unchanged, the aggregation head adds dense projections over the token state and already-computed expert outputs; a systems study should report wall-clock latency, memory, and throughput. Finally, our two backbones use Top-8 and Top-6 routed-expert selection; behavior may differ under Top-1 or Top-2 routing.

Relative to the previous draft, the main external-validity gap is substantially reduced: the diagnostic phenomenon now appears in two pretrained MoE families with different routing structures, including a shared-expert architecture. The highest-priority remaining experiments are (i) two additional DeepSeek optimization seeds with a larger locked evaluation, (ii) latency and throughput, and (iii) at least one downstream benchmark suite.

\section{Conclusion}
We separate two decisions that sparse MoE routers usually conflate: which experts to execute and how strongly to commit to each selected output. On OLMoE, fixed-dispatch counterfactuals expose substantial aggregation headroom, native router scores poorly rank within-set utility, and downstream gradients reveal useful commitment credit. A small aggregation head trained directly with the LM objective then learns effective token-adaptive commitment weights, with three-seed gains on fresh and frozen-confirmatory evaluations across WikiText, C4, and PTB.

A second-backbone audit on DeepSeek-V2-Lite shows that the diagnostic phenomenon is not specific to OLMoE: significant best-vertex headroom persists under Top-6 routed experts plus shared experts, while the router's Top1 selected expert is only chance-level at identifying the best commitment vertex. In a one-seed learned replication, \FDAA significantly improves locked WikiText and PTB and beats learned static calibration on both, while C4 remains neutral. The combined evidence therefore supports a cross-architecture separation between expert selection and expert commitment, while also showing that the amount of exploitable headroom and the transferability of a learned commitment policy are architecture- and domain-dependent. Expert selection and expert commitment are distinct optimization problems, and the latter can often be improved without changing which experts run.

\appendix
\section{Detailed Diagnostic Results}
\subsection{Three-seed fixed-dispatch oracle}
The full-horizon structured-oracle CE deltas are $-0.02015$, $-0.01234$, and $-0.01548$ for seeds 1337, 2027, and 4049. All three bootstrap intervals exclude zero. The action family contains 20 structured points and should be interpreted as a lower bound on the continuous-simplex oracle, not a global optimum.

\subsection{Gradient utility}
On 256 locked examples, the full-horizon exact vertex oracle obtains $\Delta\CE=-0.02022$ with 95\% CI $[-0.02542,-0.01619]$. Gradient argmax obtains $-0.01073$ with CI $[-0.01587,-0.00631]$, mean within-set Spearman 0.595, pairwise accuracy 0.749, Top1 exact-best accuracy 0.504, and Top2 recall 0.734. A development-selected threshold policy chooses no action on 78.1\% of tokens, has harm fraction 2.34\%, and obtains $-0.00908$ with CI $[-0.01382,-0.00537]$.

\subsection{Post-hoc predictor failures}
The exact-CF reranker uses 512 training examples, 128 validation examples, and 256 locked examples. Its raw argmax policy obtains $\Delta\CE=+0.00275$ and utility Spearman 0.036; confidence selection chooses 100\% no-op. A task-aware subspace audit constructs the Gram geometry of the eight expert displacement vectors. Pseudoinverse and ridge reconstructions achieve essentially perfect ranking agreement with gradient utility, proving that the chosen 8-D representation does not itself discard teacher decision information. Nevertheless, locked predicted-vs-gradient Spearman is $-0.015$ (local), 0.089 (post-compute), 0.044 (prefix-16), 0.098 (future-16 leakage), and 0.008 (full-future leakage). Raw exact policies are harmful and confidence selection again collapses to no-op.

\subsection{DeepSeek-V2-Lite replication sanity and selection}
At the audited layer, DeepSeek-V2-Lite uses 64 routed experts with Top-6 selection and two shared experts. The shared-expert branch is unchanged. Zero initialization gives maximum logit difference 0, CE difference 0, routed-mass error 0, and selected-ID mismatch 0. A real backward sanity check gives gradient norm 0.0677 with 65 nonzero gradient elements. For the one-seed mixed-domain run, checkpoint selection uses only the equal-domain mean normalized CE delta on 128 non-overlapping development samples per domain; harm fraction is diagnostic and is not used for selection. The selected adaptive checkpoint is step 700, while the selected static checkpoint is step 500.

\section{Pilot Tiny-Model Results}
Before moving to a pretrained MoE, we audited a 1.89M-parameter TinyStories MoE. Full-horizon structured headroom was statistically nonzero but very small ($\Delta\CE=-9.50\times10^{-5}$ on 1,024 audit examples) and concentrated in a small fraction of tokens. Detector experiments on the tiny model were not predictive enough to justify scaling that branch. We include this pilot only to document the experimental path; the paper's primary claims are based on pretrained OLMoE.

\section{Reproducibility Protocol}
\paragraph{Numerical isolation.} A recurring implementation issue in sparse MoE counterfactual evaluation is numerical coupling when multiple alternatives are placed in the same batch: expert regrouping changes matrix-multiplication shapes and can create small but consequential discrepancies. Our exact audits therefore run the native baseline, each counterfactual vertex, and each gradient pass independently with batch size 1. We require zero baseline CE mismatch and zero target expert-output mismatch before accepting a run.

\paragraph{Mass preservation.} For learned aggregation, Eq.~6 renormalizes within the native selected set and multiplies by the original total selected mass. Reported mass error is at floating-point roundoff scale (e.g., $5.96\times10^{-8}$ in Stage-1 sanity checks).

\paragraph{Checkpoint selection.} Development and locked/final splits are physically separated. Static calibrators are optimized separately from token-adaptive heads. Final WikiText-103 test and C4 validation are not used to choose the reported mixed-domain checkpoints. PTB is completely unseen during training and checkpoint selection.

\paragraph{Confirmatory evaluation.} The mixed-domain checkpoints for seeds 1337, 2027, and 4049 are frozen before confirmatory evaluation. Each of WikiText-103 test, C4 validation, and PTB test uses 1,024 fixed non-overlapping length-96 windows, evaluated independently with batch size 1. The same windows are used across optimization seeds. The primary cross-seed interval averages seed deltas within each window and bootstraps windows with 10,000 resamples, avoiding pseudo-replication from treating seed$\times$window cells as independent.

\section{Current Results Registry}
The experiment registry accompanying the project contains machine-readable quantitative results, including negative predictor experiments, the frozen three-seed OLMoE confirmatory evaluation, and the DeepSeek-V2-Lite fixed-dispatch replication. It keeps dataset splits, checkpoint hashes, evaluation windows, sanity checks, and paired-control statistics explicit so that additional DeepSeek seeds, systems measurements, and downstream experiments can be incorporated without changing the evaluation protocol.

\end{document}